\pdfoutput=1

\documentclass[11pt]{article}

\usepackage[final]{EMNLP2022}

\usepackage{xcolor}
\usepackage{soul}
\usepackage{times}
\usepackage{latexsym}
\usepackage{arydshln}
\usepackage{comment}
\usepackage{framed}

\usepackage[T1]{fontenc}

\definecolor{lightgray}{rgb}{0.83, 0.83, 0.83}

\definecolor{dollarbill}{rgb}{0.52, 0.73, 0.4}

\usepackage[utf8]{inputenc}

\usepackage{microtype}

\usepackage{inconsolata}
\definecolor{ao}{rgb}{0.0, 0.5, 0.0}
\newcommand{\cornelia}[1]{\textcolor{red}{\bf\small [#1 --cc]}}

\newcommand{\rom}[1]{\expandafter{\romannumeral #1\relax}}
\newcommand*\samethanks[1][\value{footnote}]{\footnotemark[#1]}

\title{\textsc{KPDrop}: Improving Absent Keyphrase Generation}

\author{\textbf{Jishnu Ray Chowdhury}$^\spadesuit$ $\quad$
        \textbf{Seoyeon Park}$^\spadesuit$ \thanks{$\quad$ Equal contribution} $\quad$
        \textbf{Tuhin Kundu}$^\clubsuit$ \samethanks $\;$ \thanks{$\quad$ Work done at the University of Illinois at Chicago before Amazon} $\quad$
        \textbf{Cornelia Caragea}$^\spadesuit$\\
  $^\spadesuit$ Computer Science, University of Illinois at Chicago $\quad$ $^\clubsuit$Amazon\\
  {\tt \{jraych2,spark313,cornelia\}@uic.edu} \\
  {\tt tuhinkundu@outlook.com} \\
}

\begin{document}
\maketitle
\begin{abstract}
Keyphrase generation is the task of generating phrases (keyphrases) that summarize the main topics of a given document. Keyphrases can be either present or absent from the given document. While the extraction of present keyphrases has received much attention in the past, only recently a stronger focus has been placed on the generation of absent keyphrases. However, generating absent keyphrases is challenging; even the best methods show only a modest degree of success. In this paper, we propose a model-agnostic approach called keyphrase dropout (or \textsc{KPDrop}) to improve {\em absent keyphrase generation}. In this approach, we randomly drop present keyphrases from the document and turn them into artificial absent keyphrases during training. We test our approach extensively and show that it consistently improves the absent performance of strong baselines in both supervised and resource-constrained semi-supervised settings\footnote{Code: \url{https://github.com/JRC1995/KPDrop}}.
\end{abstract}

\section{Introduction}
\label{introduction}
Keyphrase generation (KG) is the task of producing a set of phrases that best summarize a document. It can be leveraged for various applications such as text summarization \cite{zhang2017mike}, recommendation \cite{10.1145/3269206.3271696}, and opinion mining \cite{meng2012entity}. Accurate identification of keyphrases especially on scientific papers can improve efficiency in paper indexing and paper retrieval \cite{chen-etal-2019-integrated, boudin-etal-2020-keyphrase}.

Keyphrases can be divided into two types: (1) \textit{present} keyphrases and (2) \textit{absent} keyphrases. Present keyphrases appear verbatim in the document, whereas absent keyphrases are topically-relevant but missing from the document. Many prior works \cite{hasan-ng-2014-automatic, unlu2019survey} focused on present keyphrase \textit{extraction} exclusively. 
As such, they are not suitable for generating any absent keyphrases. To overcome this limitation, recent approaches \cite{meng2017deep,yuan2020one,chen-etal-2020-exclusive, ye-etal-2021-one2set} use sequence-to-sequence (seq2seq) models to generate both present and absent keyphrases. 

As shown by \citet{boudin-gallina-2021-redefining}, absent keyphrases that are substantially different from the present ones can significantly improve the effectiveness of document-retrieval. \citet{boudin-gallina-2021-redefining} suggested that absent keyphrases can improve document-retrieval by expanding the query terms to alleviate the vocabulary mismatch problem between the query terms and relevant documents \cite{furnas1987vocabulary}. Thus, there is a strong motivation for improving absent keyphrase performance. However, generating absent keyphrases can be very challenging. Even the best keyphrase generation models \cite{chen-etal-2020-exclusive, ye2021heterogeneous, ye-etal-2021-one2set} still only achieve a modest performance in absent keyphrases. 
 
\colorlet{c1}{blue!30!white}
\colorlet{c2}{red!30!white}
\colorlet{c3}{purple!30!white}
\colorlet{c4}{dollarbill!60}
\colorlet{c5}{yellow!40!white}
\colorlet{c6}{orange!30!white}

\begin{figure*}[ht]
\begin{framed}
\small
\vspace{-1mm}
\textbf{Original Input Before Applying \textsc{KPDrop}:}\\
\fbox{\parbox{0.985\textwidth}{
\textbf{Input:} \underline{The Hearing-Aid Speech Perception Index (HASPI)} \\
This paper presents a new index for predicting \colorbox{c4}{speech intelligibility} for normal-hearing and hearing-impaired listeners. The Hearing-Aid Speech Perception Index (HASPI) is based on a model of the auditory periphery that incorporates changes due to \colorbox{c4}{hearing loss}. The index compares the envelope and temporal fine structure outputs of the \colorbox{c4}{auditory model} for a reference signal to the outputs of the model for the signal under test. The \colorbox{c4}{auditory model} for the reference signal is set for normal hearing, while the model for the test signal incorporates the peripheral \colorbox{c4}{hearing loss}.

\textbf{Keyphrases}: \colorbox{c4}{Speech intelligibility}; \colorbox{c4}{Auditory model}; \colorbox{c4}{Hearing loss}; \colorbox{c1}{Hearing aids}; \colorbox{c1}{Intelligibility index}
}}\\

\textbf{New Input After Applying \textsc{KPDrop}:}\\
\fbox{\parbox{0.985\textwidth}{
\textbf{Input:} \underline{The Hearing-Aid Speech Perception Index (HASPI)} \\
This paper presents a new index for predicting \colorbox{c4}{speech intelligibility} for normal-hearing and hearing-impaired listeners. The Hearing-Aid Speech Perception Index (HASPI) is based on a model of the auditory periphery that incorporates changes due to \colorbox{c4}{hearing loss}. The index compares the envelope and temporal fine structure outputs of the \colorbox{c2}{\texttt{[MASK]}} for a reference signal to the outputs of the model for the signal under test. The \colorbox{c2}{\texttt{[MASK]}} for the reference signal is set for normal hearing, while the model for the test signal incorporates the peripheral \colorbox{c4}{hearing loss}.\\ 
\textbf{Keyphrases}: \colorbox{c4}{Speech intelligibility}; \colorbox{c1}{Auditory model}; \colorbox{c4}{Hearing loss}; \colorbox{c1}{Hearing aids}; \colorbox{c1}{Intelligibility index}\\[-.6em]}}
\end{framed}
\vspace{-2mm}
\caption{An example of how the input document \cite{KATES201475}, and the keyphrases change after applying \textsc{KPDrop}. 
Here, \textsc{KPDrop} drops the present keyphrase ``auditory model". \colorbox{c4}{Green} highlighting indicates present keyphrases, \colorbox{c1}{blue} highlighting indicates absent keyphrases, and \colorbox{c2}{red} highlighting indicates mask tokens.}
\label{fig:example}
\end{figure*}

In this work, we propose {\em keyphrase dropout} or \textsc{KPDrop} as a simple and effective technique to improve the performance of \textit{absent} keyphrases. Unlike the traditional dropout method that focuses on dropping neurons  \cite{JMLR:v15:srivastava14a}, we propose a novel dropout method where, instead of neurons, we randomly drop entire phrases (specifically, present gold keyphrases) 
from a given document during training. Thus, the dropped present keyphrases are turned into (artificial) absent keyphrases. As a result, \textsc{KPDrop} has the following two effects:
\begin{enumerate}
    \item The model is forced to deeply utilize the context information to infer dropped keyphrases that could have been otherwise simply extracted from the text. Thereby, the capability to infer missing keyphrases in general (including naturally missing ones) is increased.
    \item The method stands as a dynamic data augmentation strategy. \textsc{KPDrop} can be used to modify a given set of training documents by randomly dropping some present keyphrases from the documents turning them into artificial absent keyphrases. As such, new data can be created from the originals just like a standard data augmentation technique. Moreover, the process can be dynamically done in real-time during training. Thus, in different epochs different present keyphrases may be dropped for the same document yielding higher diversity in the data for model training.
\end{enumerate}

\noindent We apply \textsc{KPDrop} into three distinct neural models representing three distinct paradigms: one2many \cite{yuan2020one}, one2one \cite{meng2017deep, Huang_Xu_Jiao_Zu_Zhang_2021}, and one2set \cite{ye-etal-2021-one2set}. 
We observe consistent and substantial improvement in absent keyphrase performance in supervised settings on five standard datasets used for KG evaluation, with little to no drops in present keyphrase performance (and in fact often yielding improved present performance). Additionally, \textsc{KPDrop} can be used to create synthetic absent keyphrases for unlabelled data to be used for self-supervised pre-training. We demonstrate that such pre-training augmented with \textsc{KPDrop} brings substantial improvement when fine-tuned in low-resource labelled data. 


\setlength{\parskip}{0.0em}
\section{Methodology of \textsc{KPDrop}}
\label{sec:method}
In this section, we formally describe our approach of Keyphrase Dropout (\textsc{KPDrop}). In keyphrase generation (KG), we have a document $X$ as an input sequence of tokens, and a set of $n$ keyphrases, $Y = \{y_1,y_2,\dots,y_n\}$, as the target output. Within $Y$, there is a subset of $s$ ($0 \leq s \leq n$) present keyphrases as $P = \{p_1, p_2, p_3,\dots,p_s\}$ ($P \subseteq Y$) and a subset of $t$ ($0 \leq t \leq n$) absent keyphrases as $A = \{a_1, a_2, a_3,\dots,a_t\}$ ($A \subseteq Y$). It also holds true that $A \cap P = \emptyset$ and $s+t=n$. Similar to \citet{meng2017deep}, we separate absent and present keyphrases by checking if the stemmed version of a keyphrase appears in the stemmed version of the input document (in which case it is present) or does not appear (in which case it is absent).

When applying \textsc{KPDrop}, any present keyphrase $p_k \in P$ can be randomly dropped with a probability of $r$ (sampled from binomial distribution), where $r$ is the dropout rate (or \textsc{KPDrop} rate). Let $O$ ($O \subseteq P$) be the subset of present keyphrases that are randomly dropped during the application of \textsc{KPDrop} at some training epoch. For every keyphrase $p_k \in O$, we remove \textit{any and all} substrings  from $X$ that, when stemmed, match the stemmed version of the phrase $p_k$ and replace each removed substring with a special mask token \verb+[MASK]+. We call the modified version of $X$ as $X^{new}$. 
The sets of present and absent keyphrases are also modified. The new set of present keyphrases becomes $P := P \setminus O$ and the new set of absent keyphrases becomes $A := A \cup O$. Thus, the keyphrases in $O$, which were originally present, become artificially absent after applying \textsc{KPDrop}. The set $Y$ becomes $Y^{new}$ with the new versions of $P$ and $A$. An example of applying \textsc{KPDrop} is shown in Figure \ref{fig:example}. 

Note that although as a set $Y^{new}$ is the same as $Y$, we use $Y^{new}$ to convey that the sets of present and absent keyphrases have changed. 



\subsection{Two Strategies of Applying \textsc{KPDrop}}
\label{strategies}
\textsc{KPDrop} can be applied to a given mini-batch of examples in at least two distinct ways which act as data augmentation (noising) strategies. 
\setlength{\parskip}{0.5em}

\noindent {\bf \textsc{KPDrop-R}:} For the first strategy that we refer to as \textsc{KPDrop-Replace} or \textsc{KPDrop-R}, we can think of \textsc{KPDrop} as a dropout technique and like other applications of dropout we can choose to \textit{replace} the original examples with their corresponding keyphrase-dropped versions. In other words, each original sample $(X,Y)$ in a mini-batch $B$ is replaced with $(X^{new},Y^{new})$ obtained after applying \textsc{KPDrop}. More formally, if initially we had a mini-batch set $B$, after applying \textsc{KPDrop-R}, we get a new batch $B_{KPDR}$ as $B_{KPDR} = \{(X^{new},Y^{new}) | (X,Y) \in B\}$. Note that at any given training iteration, we create a \textit{single} KPDropped counterpart $(X^{new}, Y^{new})$ for each sample $(X,Y)$ in the mini-batch. 

\noindent {\bf \textsc{KPDrop-A}:} For the second strategy that we refer to as \textsc{KPDrop-Append} or \textsc{KPDrop-A}, we can think of \textsc{KPDrop} as closer to a data augmentation technique where instead of replacing the original examples we \textit{augment} the original batch with the new keyphrase dropped versions of the originals. In other words, for each original sample $(X,Y)$ in a mini-batch, we add $(X^{new},Y^{new})$ obtained after applying \textsc{KPDrop}. More formally, starting with the same mini-batch set $B$ as before, after applying \textsc{KPDrop-A}, we get a new batch $B_{KPDA}$ as $B_{KPDA} = B \cup B_{KPDR}$.

\noindent Given that \textsc{KPDrop-R} can increase the number of absent keyphrases per sample during training, we hypothesize that  the model can learn to be more biased towards generating absent keyphrases. This can help to increase absent keyphrase performance at the cost of present keyphrase performance because the technique will be dropping present keyphrases. 
On the other hand, \textsc{KPDrop-A}, instead of replacing the original data, it can offer extra samples per batch that have additional artificially absent keyphrases. Thus, \textsc{KPDrop-A} should be still able to improve absent keyphrase performance while also maintaining the original data with its original present keyphrases. Thus, \textsc{KPDrop-A} will offer the underlying model with the same opportunity to learn present keyphrases as would the model without \textsc{KPDrop-A}. Intuitively, this can help improve absent performance without a substantial cost to present performance. 
\setlength{\parskip}{0.0em}



\subsection{\textsc{KPDrop} Features and Connections to Works from Other Areas}
In this section, we highlight some notable features of \textsc{KPDrop} and draw connections to some relevant works outside the area of keyphrase generation. 
\setlength{\parskip}{0.5em}

\noindent \textbf{\textsc{KPDrop} and Masked Language Modeling:} Superficially, \textsc{KPDrop} is similar to a standard masked language modeling (MLM) \cite{devlin2018bert, raffel2020exploring} strategy insofar that both involve the reconstruction of some masked out text-spans. However, there are several crucial differences between \textsc{KPDrop} and vanilla MLM. First, \textsc{KPDrop} masks only \textit{present keyphrases}, not random subspans or phrases. Second, \textsc{KPDrop} masks \textit{all instances} of the selected present keyphrases in the document to make them truly absent (see Figure \ref{fig:example}). In contrast, MLM does not mask \textit{all instances} of the masked token or phrase. Third, masking (replacing a subspan with a mask token) is only an engineering choice for \textsc{KPDrop}; not an essential ingredient. We can simply drop the selected present keyphrases without replacing them with a mask token. 
In \textsc{KPDrop}, predictions of artificial absent keyphrases (masked present keyphrases) are not \emph{explicitly} associated with any mask position.

\vspace{1mm}
\noindent \textbf{\textsc{KPDrop} as Structured Dropout:} \textsc{KPDrop} is a unique kind of \textit{structured dropout} where all instances of some randomly chosen present keyphrases are dropped. There are other examples of structured dropout in prior works. For example, \citet{iyyer-etal-2015-deep} proposed word dropout (for general NLP tasks) where whole word embeddings are dropped instead of just random neurons. \citet{huang206deep} and \citet{Fan2020Reducing} used another form of structured dropout that stochastically drops whole layers for general computer vision tasks and for general NLP tasks respectively. 


\vspace{1mm}
\noindent \textbf{Model-Agnosticism of \textsc{KPDrop}:} One important feature of \textsc{KPDrop} is that it is model-agnostic. \textsc{KPDrop} only requires changing the input and output without making internal architectural changes. As such, it is compatible with any model that is suitable for KG. \textsc{KPDrop} can also be easily stacked with other techniques that help absent keyphrase generation. Later (in $\S$\ref{absent_results}), we demonstrate that \textsc{KPDrop} works just as well for very different architectures: (1) an RNN-based model trained in one2many settings \cite{yuan2020one} (2) an RNN-based model trained in one2one settings \cite{meng2017deep, Huang_Xu_Jiao_Zu_Zhang_2021}, and (3) a Transformer-based model trained in one2set settings \cite{ye-etal-2021-one2set}.



\setlength{\parskip}{0.0em}

\begin{table*}[t]
\centering
\small
\def\arraystretch{1.2}
\begin{tabular}{l|cc|cc|cc|cc|cc}
\hline
 & \multicolumn{2}{c}{\textbf{Inspec}} & \multicolumn{2}{|c}{\textbf{NUS}} & \multicolumn{2}{|c}{\textbf{Krapivin}} & \multicolumn{2}{|c}{\textbf{SemEval}} & \multicolumn{2}{|c}{\textbf{KP20k}} \\  
\multicolumn{1}{c|}{\textbf{Models}} & \textbf{F1@M} & {\textbf{F1@5}}  & \textbf{F1@M} & {\textbf{F1@5}} & \textbf{F1@M} & {\textbf{F1@5}} & \textbf{F1@M} & {\textbf{F1@5}} & \textbf{F1@M} & \textbf{F1@5} \\ \hline
\multicolumn{11}{l}{\textbf{GRU One2Many}}\\ \hline
Greedy & $1.4_2$ & $1.4_2$ & $2.9_6$ & $2.9_6$ & $3.4_4$ & $3.4_4$ & $2.4_2$ & $2.4_2$ & $3.3_3$ & $3.3_3$\\
Greedy+\textsc{KPD-R} & $1.2_1$ & $1.2_1$ & $3.2_7$ & $3.2_7$ & $\mathbf{5.1_2}$ & $\mathbf{5.1_2}$ & $\mathbf{2.7_2}$ & $\mathbf{2.7_2}$ & $\mathbf{4.0_1}$ & $\mathbf{4.0_1}$\\
Greedy+\textsc{KPD-A} & $\mathbf{1.5_1}$ & $\mathbf{1.5_1}$ & $\mathbf{3.5_6}$ & $\mathbf{3.5_6}$ & $4.6_3$ & $4.6_3$ & $2.6_4$ & $2.6_4$ & $3.9_3$ & $3.9_3$ \\ \hdashline
Beam & $2.8_1$ & $\mathbf{2.9_1}$ & $5.5_3$ & $5.5_2$ & $7.2_3$ & $7.3_4$ & $3.8_2$ & $3.8_2$ & $5.8_0$ & $5.8_1$\\
Beam+\textsc{KPD-R} & $2.8_3$ & $\mathbf{2.9_3}$ & $6.5_2$ & $\mathbf{6.5_2}$ & $\mathbf{7.3_3}$ & $\mathbf{7.4_3}$ & $\mathbf{5.1_2}$ & $4.8_2$ & $6.1_0$ & $6.3_0$\\
Beam+\textsc{KPD-A} & $\mathbf{3.0_1}$ & $\mathbf{2.9_1}$ & $\mathbf{6.6_{10}}$ & $\mathbf{6.5_9}$ & $\mathbf{7.3_3}$ & $7.3_4$ & $5.0_3$ & $\mathbf{5.0_4}$ & $\mathbf{6.4_0}$ & $\mathbf{6.5_0}$\\ 
\hline
\multicolumn{11}{l}{\textbf{GRU One2One}}\\ \hline
Beam & $0.6_0$ & $2.8_1$ & $1.5_1$ & $5.7_7$ & $1.2_0$ & $5.9_3$ & $1.3_0$ & $3.8_1$ & $1.0_0$ & $6.2_0$ \\
Beam+\textsc{KPD-R} & $\mathbf{0.7_0}$ & $\mathbf{2.9_3}$ & $1.5_0$ & $\mathbf{7.5_3}$ & $\mathbf{1.3_0}$ & $\mathbf{7.8_2}$ & $1.4_0$ & $\mathbf{4.9_5}$ & $1.0_0$ & $6.5_1$\\
Beam+\textsc{KPD-A} & $\mathbf{0.7_0}$ & $2.7_1$ & $\mathbf{1.6_0}$ & $6.7_2$ & $\mathbf{1.3_0}$ & $6.5_3$ & $\mathbf{1.5_1}$ & $4.1_3$ & $\mathbf{1.1_0}$ & $\mathbf{6.8_0}$\\
\hline
\multicolumn{11}{l}{\textbf{Transformer One2Set}}\\ \hline
Greedy & $2.8_4$ & $2.8_4$ & $6.4_8$ & $6.4_8$ & $6.8_2$ & $6.8_2$ & $3.5_1$ & $3.5_1$ & $5.6_0$ & $5.5_0$\\
Greedy+\textsc{KPD-R} & $2.9_1$ & $2.8_1$ & $6.9_8$ & $6.9_7$ & $\mathbf{8.4_8}$ & $\mathbf{8.4_8}$ & $4.6_3$ & $\mathbf{4.6_4}$ & $6.4_1$ & $6.3_1$\\
Greedy+\textsc{KPD-A} & $\mathbf{3.2_2}$ & $\mathbf{3.2_2}$ & $\mathbf{7.4_9}$ & $\mathbf{7.4_9}$ & $7.2_7$ & $7.2_7$ & $\mathbf{4.7_1}$ & $\mathbf{4.6_1}$ & $\mathbf{6.6_1}$ & $\mathbf{6.5_1}$\\\hdashline
Beam & $0.4_0$ & $3.3_2$ & $0.9_0$ & $7.0_5$ & $0.8_0$ & $6.7_5$ & $0.8_0$ & $4.7_4$ & $\mathbf{0.6_0}$ & $5.8_0$\\
Beam+\textsc{KPD-R} & $0.4_0$ & $2.4_1$ & $\mathbf{1.0_0}$ & $7.2_5$ & $\mathbf{0.9_0}$ & $7.3_2$ & $0.9_0$ & $5.2_4$ & $\mathbf{0.6_0}$ & $6.1_0$\\
Beam+\textsc{KPD-A} & $\mathbf{0.5_0}$ & $\mathbf{3.6_2}$ & $\mathbf{1.0_0}$ & $\mathbf{7.9_4}$ & $\mathbf{0.9_0}$ & $\mathbf{7.8_2}$ & $\mathbf{1.0_0}$ & $\mathbf{5.3_4}$ & $\mathbf{0.6_0}$ & $\mathbf{6.7_0}$\\ 
\hline
\end{tabular}
\caption{Absent keyphrase performance (\textbf{F1}) for different models. KPD represents KPDrop. Subscripts represent standard deviation (e.g., $31.1_1$ represents $31.1\pm0.1$). We bold the best scores per block.} 
\label{tb:absent_result}
\end{table*}

\begin{table*}[t]
\centering
\small
\def\arraystretch{1.2}
\begin{tabular}{l|cc|cc|cc|cc|cc}
\hline
 & \multicolumn{2}{c}{\textbf{Inspec}} & \multicolumn{2}{|c}{\textbf{NUS}} & \multicolumn{2}{|c}{\textbf{Krapivin}} & \multicolumn{2}{|c}{\textbf{SemEval}} & \multicolumn{2}{|c}{\textbf{KP20k}} \\  
\multicolumn{1}{c|}{\textbf{Models}} & \textbf{R@10} & {\textbf{R@50}}  & \textbf{R@10} & {\textbf{R@50}} & \textbf{R@10} & {\textbf{R@50}} & \textbf{R@10} & {\textbf{R@50}} & \textbf{R@10} & \textbf{R@50} \\ \hline
\multicolumn{11}{l}{\textbf{GRU One2Many}}\\ \hline
Beam & $3.8_1$ & $3.8_1$ & $5.3_2$ & $5.3_2$ & $8.2_3$ & $8.2_3$ & $3.0_0$ & $3.0_0$ & $8.1_1$ & $8.1_1$\\
Beam+\textsc{KPD-R} & $\mathbf{4.3_3}$ & $\mathbf{4.3_4}$ & $\mathbf{7.5_4}$ & $\mathbf{7.6_3}$ & $\mathbf{10.9_1}$ & $\mathbf{11.1_2}$ & $\mathbf{4.2_3}$ & $\mathbf{4.2_3}$ & $\mathbf{10.1_1}$ & $\mathbf{10.2_1}$\\
Beam+\textsc{KPD-A} & $4.2_2$ & $4.2_2$ & $6.4_{10}$ & $6.4_{10}$ & $9.0_3$ & $9.1_3$ & $3.9_2$ & $4.0_2$ & $9.2_1$ & $9.3_1$\\ 
\hline
\multicolumn{11}{l}{\textbf{GRU One2One}}\\ \hline
Beam & $6.1_3$ & $11.3_2$ & $9.5_{10}$ & $17.4_5$ & $12.3_2$ & $22.9_1$ & $4.4_2$ & $8.1_7$ & $14.0_0$ & $23.5_1$\\
Beam+\textsc{KPD-R} & $\mathbf{6.3_3}$ & $12.2_4$ & $\mathbf{12.1_9}$ & $\mathbf{20.3_4}$ & $\mathbf{15.6_9}$ & $\mathbf{26.8_6}$ & $\mathbf{5.6_3}$ & $\mathbf{9.4_3}$ & $14.6_1$ & $24.8_3$\\
Beam+\textsc{KPD-A} & $5.9_4$ & $\mathbf{12.9_7}$ & $11.3_7$ & $19.0_5$ & $14.5_7$ & $24.7_5$ & $5.0_4$ & $9.3_3$ & $\mathbf{15.2_1}$ & $\mathbf{25.5_1}$ \\
\hline
\multicolumn{11}{l}{\textbf{Transformer One2Set}}\\ \hline
Beam & $6.7_3$ & $12.5_3$ & $12.0_5$ & $19.6_3$ & $13.6_{10}$ & $24.4_9$ & $5.4_2$ & $9.1_2$ & $13.5_2$ & $23.4_2$\\
Beam+\textsc{KPD-R} & $5.7_1$ & $11.7_3$ & $13.3_6$ & $21.2_5$ & $\mathbf{15.6_5}$ & $27.3_9$ & $\mathbf{5.9_8}$ & $10.4_5$ & $14.5_0$ & $25.4_2$\\
Beam+\textsc{KPD-A} & $\mathbf{7.9_3}$ & $\mathbf{13.6_4}$ & $\mathbf{14.0_2}$ & $\mathbf{22.3_8}$ & $15.2_4$ & $\mathbf{27.7_4}$ & $\mathbf{5.9_5}$ & $\mathbf{11.0_7}$ & $\mathbf{15.6_2}$ & $\mathbf{26.6_2}$\\ 
\hline
\end{tabular}
\caption{Absent keyphrase performance (\textbf{Recall}) for different models. KPD represents KPDrop. Subscripts represent standard deviation (e.g., $31.1_1$ represents $31.1\pm0.1$). We bold the best scores per block.} 
\label{tb:absent_result_recall}
\end{table*}

\begin{table*}[t]
\centering
\small
\def\arraystretch{1.2}
\begin{tabular}{l|cc|cc|cc|cc|cc}
\hline
 & \multicolumn{2}{c}{\textbf{Inspec}} & \multicolumn{2}{|c}{\textbf{NUS}} & \multicolumn{2}{|c}{\textbf{Krapivin}} & \multicolumn{2}{|c}{\textbf{SemEval}} & \multicolumn{2}{|c}{\textbf{KP20k}} \\  
\multicolumn{1}{c|}{\textbf{Models}} & \textbf{F1@M} & {\textbf{F1@5}}  & \textbf{F1@M} & {\textbf{F1@5}} & \textbf{F1@M} & {\textbf{F1@5}} & \textbf{F1@M} & {\textbf{F1@5}} & \textbf{F1@M} & \textbf{F1@5} \\ \hline
\multicolumn{11}{l}{\textbf{GRU One2Many}}\\ \hline
Greedy & $\mathbf{27.4_7}$ & $\mathbf{27.1_8}$ & $38.1_3$ & $37.7_6$ & $\mathbf{34.7_9}$ & $\mathbf{34.2_7}$ & $\mathbf{29.6_{10}}$ & $\mathbf{29.1_9}$ & $\mathbf{37.3_1}$ & $\mathbf{37.0_0}$\\
Greedy+\textsc{KPD-R} & $18.7_3$ & $18.7_3$ & $28.1_3$ & $28.1_3$ & $28.4_6$ & $28.4_6$ & $20.1_{14}$ & $20.1_{14}$ & $30.8_2$ & $30.8_2$\\
Greedy+\textsc{KPD-A} & $25.1_4$ & $24.8_3$ & $\mathbf{38.5_4}$ & $\mathbf{38.2_3}$ & $34.1_9$ & $33.5_8$ & $27.9_7$ & $27.7_6$ & $37.0_2$ & $36.7_2$\\\hdashline
Beam & $38.7_3$ & $34.4_6$ & $37.0_6$ & $41.6_5$ & $25.6_5$ & $32.8_3$ & $31.9_{12}$ & $33.4_3$ & $32.2_1$ & $36.8_0$ \\
Beam+\textsc{KPD-R} & $34.9_2$ & $33.2_2$ & $\mathbf{42.0_4}$ & $41.0_2$ & $\mathbf{36.7_3}$ & $\mathbf{36.7_2}$ & $\mathbf{34.6_9}$ & $\mathbf{34.1_9}$ & $\mathbf{36.3_1}$ & $36.5_1$\\
Beam+\textsc{KPD-A} & $\mathbf{38.9_5}$ & $\mathbf{34.6_5}$ & $39.4_5$ & $\mathbf{41.8_7}$ & $28.4_0$ & $34.2_8$ & $33.0_{17}$ & $33.2_7$ & $33.9_2$ & $\mathbf{37.1_1}$\\ 
\hline
\multicolumn{11}{l}{\textbf{GRU One2One}}\\ \hline
Beam & $25.5_1$ & $\mathbf{30.6_1}$ & $16.5_3$ & $40.2_{10}$ & $12.4_1$ & $29.8_1$ & $16.5_2$ & $29.2_{16}$ & $13.1_0$ & $39.4_1$\\
Beam+\textsc{KPD-R} & $\mathbf{29.6_3}$ & $30.4_3$ & $\mathbf{24.6_4}$ & $4\mathbf{3.9_3}$ & $\mathbf{17.5_1}$ & $\mathbf{36.5_2}$ & $\mathbf{23.6_4}$ & $\mathbf{32.7_{10}}$ & $\mathbf{17.0_2}$ & $37.9_0$\\
Beam+\textsc{KPD-A} & $26.3_4$ & $\mathbf{30.6_2}$ & $17.1_0$ & $38.9_6$ & $12.9_0$ & $29.4_4$ & $17.2_5$ & $30.0_{11}$ & $13.9_1$ & $\mathbf{39.6_0}$\\
\hline
\multicolumn{11}{l}{\textbf{Transformer One2Set}}\\ \hline
Greedy & $\mathbf{32.2_7}$ & $\mathbf{31.3_6}$ & $43.7_0$ & $42.1_3$ & $35.2_5$ & $\mathbf{34.3_{11}}$ & $\mathbf{34.7_2}$ & $33.4_7$ & $39.2_1$ & $37.9_4$\\
Greedy+\textsc{KPD-R} & $21.1_3$ & $21.0_3$ & $35.0_9$ & $34.9_{12}$ & $34.2_7$ & $34.2_7$ & $26.9_7$ & $27.0_8$ & $34.9_2$ & $34.8_2$ \\
Greedy+\textsc{KPD-A} & $30.6_3$ & $29.8_4$ & $\mathbf{44.4_3}$ & $\mathbf{42.6_3}$ & $\mathbf{35.3_6}$ & $34.0_6$ & $34.4_5$ & $\mathbf{33.6_5}$ & $\mathbf{39.6_2}$ & $\mathbf{38.5_0}$\\\hdashline
Beam & $21.7_0$ & $\mathbf{32.3_4}$ & $15.5_1$ & $\mathbf{42.3_3}$ & $11.0_1$ & $32.9_6$ & $16.3_2$ & $33.5_8$ & $10.9_0$ & $\mathbf{36.4_5}$\\
Beam+\textsc{KPD-R} & $\mathbf{23.2_2}$ & $27.8_8$ & $\mathbf{21.0_5}$ & $40.0_9$ & $\mathbf{14.8_8}$ & $\mathbf{33.6_7}$ & $\mathbf{20.9_5}$ & $32.0_5$ & $\mathbf{15.0_3}$ & $35.1_3$\\
Beam+\textsc{KPD-A} & $22.7_1$ & $32.2_6$ & $16.8_2$ & $41.8_3$ & $11.6_2$ & $32.3_7$ & $17.6_4$ & $\mathbf{34.3_{12}}$ & $11.7_2$ & $36.3_0$\\ 
\hline
\end{tabular}
\caption{Present keyphrase performance (\textbf{F1}) for different models. KPD represents KPDrop. Subscripts represent standard deviation (e.g., $31.1_1$ represents $31.1\pm0.1$). We bold the best scores per block.} 
\label{tb:present_result}
\end{table*}

\begin{table*}[t]
\centering
\small
\def\arraystretch{1.2}
\begin{tabular}{l|cc|cc|cc|cc|cc}
\hline
 & \multicolumn{2}{c}{\textbf{Inspec}} & \multicolumn{2}{|c}{\textbf{NUS}} & \multicolumn{2}{|c}{\textbf{Krapivin}} & \multicolumn{2}{|c}{\textbf{SemEval}} & \multicolumn{2}{|c}{\textbf{KP20k}} \\  
\multicolumn{1}{c|}{\textbf{Models}} & \textbf{F@M} & {\textbf{F1@5}}  & \textbf{F1@M} & {\textbf{F1@5}} & \textbf{F1@M} & {\textbf{F1@5}} & \textbf{F1@M} & {\textbf{F1@5}} & \textbf{F1@M} & \textbf{F1@5} \\ \hline
\multicolumn{11}{c}{\textbf{Absent Keyphrase Performance}}\\\hline
\multicolumn{11}{l}{Unsupervised Extraction Model \cite{liang-etal-2021-unsupervised}}\\\hline
& --- & 0.0 & --- & 0.0 & --- & 0.0 & --- & 0.0 & --- & 0.0\\
\hline
\multicolumn{11}{l}{GRU One2Many Models (Beam Search)}\\
\hline
PT & $0.0_1$ & $0.0_1$ & $0.0_0$ & $0.0_0$ & $0.0_0$ & $0.0_1$ & $0.0_1$ & $0.0_0$ & $0.0_0$ & $0.0_0$\\ 
PT+KPD-R  & $0.7_1$ & $0.7_1$ & $0.8_2$ & $0.7_1$ & $0.6_2$ & $0.5_2$ & $0.3_1$ & $0.2_2$ & $0.8_1$ & $0.8_1$\\
PT+KPD-A & $0.1_0$ & $0.1_0$ & $0.0_0$ & $0.0_0$ & $0.0_0$ & $0.0_0$ & $0.0_0$ & $0.0_0$ & $0.0_0$ & $0.0_0$\\
FT & $0.7_1$ & $0.6_1$ & $1.1_0$ & $1.1_0$ & $0.5_0$ & $0.6_0$ & $0.3_1$ & $0.3_1$ & $0.7_1$ & $0.7_1$\\
PT; FT & $0.7_2$ & $0.7_2$ & $0.5_4$ & $0.5_4$ & $0.1_0$ & $0.1_0$ & $0.1_1$ & $0.1_1$ & $0.4_1$ & $0.3_1$ \\
PT+KPD-A; FT & $0.9_2$ & $0.9_2$ & $2.5_3$ & $2.6_4$ & $1.8_2$ & $1.8_2$ & $1.8_1$ & $1.8_1$ & $1.8_1$ & $1.8_1$\\
PT+KPD-R; FT & $\mathbf{1.7_3}$ & $\mathbf{1.7_3}$ & $\mathbf{2.9_3}$ & $\mathbf{2.8_3}$ & $\mathbf{2.8_6}$ & $\mathbf{2.8_5}$ & $\mathbf{2.0_1}$ & $\mathbf{2.0_1}$ & $\mathbf{2.7_2}$ & $\mathbf{2.7_2}$\\
\hline
\multicolumn{11}{c}{\textbf{Present Keyphrase Performance}}\\\hline
\multicolumn{11}{l}{Unsupervised Extraction Model \cite{liang-etal-2021-unsupervised}}\\\hline
 & --- & $32.6$ & --- & $20.8$ & --- & $18.1$ & --- & $13.02$ & --- & $17.9$\\
\hline
\multicolumn{11}{l}{GRU One2Many Models (Beam Search)}\\
\hline
PT & $\mathbf{40.9_3}$ & $36.0_3$ & $20.6_2$ & $22.4_{10}$ & $18.5_2$ & $17.9_2$ & $22.8_5$ & $22.2_{11}$ & $16.0_1$ & $17.9_2$\\ 
PT+KPD-R & $36.2_{13}$ & $34.3_4$ & $24.8_{13}$ & $23.6_{20}$ & $19.2_1$ & $18.9_4$ & $25.1_6$ & $23.2_{11}$ & $18.6_8$ & $18.0_7$\\
PT+KPD-A & $40.2_3$ & $35.7_4$ & $21.0_6$ & $22.8_5$ & $18.8_5$ & $19.8_2$ & $23.4_6$ & $22.7_6$ & $16.1_2$ & $17.9_2$\\
FT & $27.8_{11}$ & $26.5_9$ & $32.6_2$ & $31.8_3$ & $26.6_8$ & $26.4_8$ & $27.8_8$ & $26.7_7$ & $27.1_4$ & $26.9_4$\\
PT; FT & $36.8_{17}$ & $33.5_{25}$ & $33.8_3$ & $33.5_2$ & $26.2_{12}$ & $27.5_6$ & $29.1_5$ & $27.6_7$ & $26.2_3$ & $27.2_2$ \\
PT+KPD-A; FT & $39.9_5$ & $36.0_{10}$ & $\mathbf{36.3_4}$ & $\mathbf{35.7_2}$ & $29.2_4$ & $29.6_3$ & $\mathbf{32.0_4}$ & $\mathbf{30.8_{12}}$ & $28.2_3$ & $28.8_2$\\
PT+KPD-R; FT & $40.0_{11}$ & $\mathbf{36.9_{10}}$ & $36.2_{11}$ & $34.6_9$ & $\mathbf{31.5_3}$ & $\mathbf{31.1_7}$ & $30.9_5$ & $29.9_{14}$ & $\mathbf{29.7_7}$ & $\mathbf{29.8_8}$\\
\hline
\end{tabular}%
\vspace{-2mm}
\caption{\small{Absent and present keyphrase performance using Beam Search for GRU One2Many models in a semi-supervised setup. KPD represents KPDrop. PT represents pre-training on the synthetic data (UC). PT+KPD-R or PT+KPD-A represents pre-training on UC with KPD-R or KPD-A repsectively. FT represents fine tuning or training on the low resource labelled data (LR). PT (or PT+KPD-A or PT+KPD-R) followed by ``; FT" represents that the pre-training was followed by fine-tuning of the pre-trained model on LR. We bold the overall best scores. Subscripts represent standard deviation (e.g., $31.1_1$ represents $31.1\pm0.1$).}} 
\label{tb:sm_result}
\end{table*}

\section{Evaluations}
We use the following evaluation metrics:
\setlength{\parskip}{0.5em}

\noindent \textbf{F}$_1$\textbf{@M}: F$_1$@M is a standard metric \cite{yuan2020one, chen2019guided, chan-etal-2019-neural, chen-etal-2020-exclusive, ye-etal-2021-one2set} used to evaluate the performance of keyphrase generation. This is an F$_1$ based metric where \textit{all} the predictions by a given model are considered for evaluation. 

\noindent \textbf{F}$_1$\textbf{@5}: F$_1$@5 is similar to F$_1$@M but only the top $5$ predicted keyphrases are used for evaluation (if the total number of predictions are less than $5$, all the available predictions are used). This metric is used in several prior works \cite{meng2017deep, yuan2020one}.\footnote{Note that the F$_1$@5 metric as used by us and as introduced by \citet{meng2017deep} for keyphrase generation is different from the  F$_1$@5 metric as introduced by \citet{chan-etal-2019-neural} and used in some other works \cite{chen-etal-2020-exclusive, ye-etal-2021-one2set}. We report results with \citet{chan-etal-2019-neural}'s formulation of F$_1$@5 in Appendix \ref{more_evaluation}} This metric is useful in settings where the model is used to overgenerate keyphrases (for example, by using beam search). Overgeneration can lead to lower precision when all the predictions are kept (thus, low F$_1$@M). So, often in such settings, it is more useful to check the performance of the model when some simple truncation policy is used, for example, selecting some top $k$ (e.g., top $5$ in F$_1$@5) predictions. 



\noindent \textbf{R@10 and R@50}: R@10 and R@50 represents macro recall of the top $10$ predictions and the top $50$ predictions, respectively. In some applications (such as retrieval) high recall can be sometimes more useful than precision. R@10 and R@50 are used in prior works \cite{meng2017deep, chen-etal-2018-keyphrase, liu2020keyphrase} to measure absent keyphrase performance after overgenerating them using beam search with high beam size. 

\noindent Like prior works, we use macro-F$_1$ and macro-recall for all the above metrics.  
\setlength{\parskip}{0.0em}
\section{Baselines}
\label{baselines}
We use the following baselines with our \textsc{KPDrop-R} and \textsc{KPDrop-A} approaches:
\setlength{\parskip}{0.5em}

\noindent \textbf{GRU One2Many}: GRU One2Many (also known as CatSeq) represents a simple seq2seq baseline based on GRUs. It takes a document input and generates a concatenated sequence of keyphrases similar to \citet{yuan2020one}. For this baseline, we concatenate the ground truth keyphrases based on the best performing ordering procedure \cite{meng-etal-2021-empirical}: we first concatenate the present keyphrases according to the order of their first appearance in text and then we append the absent keyphrases in their natural order. However, when using \textsc{KPDrop}, we start with the ordering as mentioned but then shift the dropped present keyphrases (artificial absent keyphrases) after the fully present keyphrases but keep them before the naturally absent keyphrases. We maintain the internal order of the dropped present (artificially absent) keyphrases. We use ``;" as the delimiter for separating keyphrases. 

\noindent \textbf{GRU One2One}: GRU One2One (also known as CopyRNN) is another seq2seq model based on GRUs. However, unlike the One2Many model, it can predict only one keyphrase per generated sequence. It can be still used to generate multiple keyphrases by using beam search and preserving multiple beams of sequences (each representing a keyphrase). The One2One approach was first introduced by \citet{meng2017deep} where the original training data was divided such that each input was associated with only one ground truth keyphrase. However, instead of that, we use the more efficient training approach of using reset states as in \citet{Huang_Xu_Jiao_Zu_Zhang_2021}. That is, we simply train the one2one model similar to one2many models using teacher forcing but we ``reset" the hidden state when generating the first word of any ground truth keyphrase. Resetting \cite{Huang_Xu_Jiao_Zu_Zhang_2021} corresponds to using the initial RNN hidden state and the first special input token indicating start of sequence and thereby removing dependencies from previous generations. 

\noindent \textbf{Transformer One2Set}: Transformer One2Set (also known as SetTrans) is a Transformer-based model trained in a new One2Set paradigm as introduced by \citet{ye-etal-2021-one2set}. In this method, a fixed number of preset control codes are used to generate all present and absent keyphrases in parallel independent of each other. Moreover, during training target keyphrases are matched with the predicted ones using Hungarian algorithm \cite{hungarian} so that the training is not sensitive to the order of the target keyphrases. Note that Transformer One2Set uses specialized control codes for present and absent keyphrase generation separately. Thus, when using \textsc{KPDrop}, we make sure to use the control codes for absent keyphrases to generate artificial absent keyphrases (dropped present keyphrases) as well. 
\setlength{\parskip}{0.0em}

\section{Supervised Experiments}
In the supervised setting, we explore the effects of applying both \textsc{KPDrop-R} and \textsc{KPDrop-A} on the baselines that we discussed in $\S$\ref{baselines}. Note that One2Many models \cite{yuan2020one, meng-etal-2021-empirical} had been explored in both greedy search based generations where only a single concatenated sequence of keyphrases is generated and also in beam search based generations where multiple beams of concatenated sequence of keyphrases are generated. We too explore both greedy and beam  search for the One2Many models. In One2One models, greedy search is ineffective because it only generates a single keyphrase. Thus, following \citet{meng2017deep}, we only use beam search to overgenerate multiple beams of keyphrases for One2One models. For One2Set models, only greedy search was explored \cite{ye-etal-2021-one2set}. Here, in addition to greedy search, we also explore using beam search for each control code in One2Set models. For all models, we investigate the effect of applying \textsc{KPDrop} in all of their applicable decoding settings (be it beam search or greedy search).
Following prior works \cite{meng2017deep, chen-etal-2018-keyphrase}, we mainly use beam search to demonstrate recall performance (Table \ref{tb:absent_result_recall}) for absent keyphrase performance. The recall performance stands out best when using beam search. Greedy search can be more conservative with generation; thus, has limited recall. We evaluate our three baselines on five scientific datasets: KP20k \cite{meng2017deep}, Inspec \cite{hulth-2003-improved}, Krapivin \cite{krapivin2009large}, SemEval \cite{kim-etal-2010-semeval} and NUS \cite{nus}. Following previous works, we use the training set of KP20k to train all models. All models are run three times on different seeds. We report the mean and standard deviation of these three runs. Hyperparameters are detailed in Appendix \ref{hyperparameters}. 

\setlength{\parskip}{0.0em}

\subsection{Results on Absent Keyphrase Generation}
\label{absent_results}
In Table \ref{tb:absent_result}, we show the F1 performance for absent keyphrase generation. As we can see from the table, \textsc{KPDrop} always boosts the absent keyphrase performance against a comparable baseline (regardless of whether we use \textsc{KPDrop-R} or \textsc{KPDrop-A}). Comparing greedy with beam search in GRU One2Many, we find that overgenerating with beam search can substantially improve the absent performance over greedy and further applying \textsc{KPDrop} to beam search improves the performance significantly (e.g., on KP20K performance improves from 5.8\% to 6.5\%). In both GRU One2One and Transformer One2Set, we observe that beam search leads to overgeneration reducing precision and thus the F1@M performance. However, F1@5 generally improves in beam search compared to greedy because only the top $5$ keyphrases are considered in this metric. Either way, whether using greedy or beam search, in both One2One models and One2Set models, \textsc{KPDrop} enhances the base performance.  Overall, for absent performance, \textsc{KPDrop-R} and \textsc{KPDrop-A} are both competitive against each other. 

In Table \ref{tb:absent_result_recall}, we show the recall performance for absent keyphrase generation when using high beam size. We find that applying any of the \textsc{KPDrop} techniques substantially improves the recall performance of absent keyphrase generation in all settings and datasets. In GRU One2Many and GRU One2One, \textsc{KPDrop-R}  generally performs better than \textsc{KPDrop-A}. Interestingly, in Transformer One2Set \textsc{KPDrop-A} generally performs better than \textsc{KPDrop-R} across all datasets. 

The above results validate our intuition that dropping keyphrases in \textsc{KPDrop} forces the models to deeply utilize the context information to learn to predict the dropped keyphrases, and thus, yields more robust models for absent keyphrase.  


\subsection{Results on Present Keyphrase Generation}
\label{present_results}
In Table \ref{tb:present_result}, we show the F1 performance for present keyphrase generation. As we hypothesized before, on greedy decoding \textsc{KPDrop-R} can significantly drop the performance of present keyphrases. However, \textsc{KPDrop-A} performs quite competitively against the baselines even in present keyphrase performance. Thus, \textsc{KPDrop-A} can be a balanced approach to boost absent performance without significantly downgrading the present performance. Interestingly, when combined with beam search, even \textsc{KPDrop-R} becomes competitive in present keyphrase generation. While, as we suggested before (in \S\ref{strategies}), \textsc{KPDrop-R} can create a tendency to undergenerate present keyphrases (given that many of them get turned artificially absent), beam search can fight against that tendency through overgeneration. This can sometimes lead to a ``sweet spot" when beam search is combined with \textsc{KPDrop-R} making it competitive even in present performance for One2Many and One2One settings. 

\section{Semi-Supervised Experiments}
We also investigate whether \textsc{KPDrop} has something to offer in a low-resource semi-supervised setting. We simulate a semi-supervised setting by randomly splitting the training set of KP20K into two parts. In one part, we keep $5000$ samples with their original keyphrases intact, but in the other part, we keep the rest of the data after removing the original keyphrases. Thus, we are left with a low-resource ($5000$ samples) author-annotated training data, and a huge unlabelled corpus (rest of the KP20K training set). Henceforth, we refer to the former low-resource labelled dataset as LR, and the latter unlabelled corpus as UC. 

We investigate a pre-training based approach to utilize UC. Essentially, we pre-train our models first on UC and then fine-tune them on LR. For pre-training, similar to \citet{ye-wang-2018-semi}, we create synthetic labels using an unsupervised keyphrase extraction model. Unlike \citet{ye-wang-2018-semi}, we use a contemporary embedding-based keyphrase extraction model \cite{liang-etal-2021-unsupervised} to generate the synthetic keyphrases. Particularly, we rank the candidate keyphrases and keep the top $10$. Note that in this pre-training setup, the synthetic keyphrases will only be present keyphrases because they are extracted from the input text. This is where \textsc{KPDrop} can make the unsupervised pre-training more interesting by creating artificial absent keyphrases by dropping of the synthetic present keyphrases (and simulating real data). We hypothesize that the application of \textsc{KPDrop} can help our models to learn more effective weights from UC in the self-supervised pre-training stage. We test the effectiveness of using \textsc{KPDrop} to augment self-supervised pre-training by testing the pre-trained model on the labelled test sets after fine-tuning on LR.  During fine-tuning on LR, for all models, we always use \textsc{KPDrop-A} as we have already shown this to be beneficial in most supervised contexts. Hyperparameters are detailed in the Appendix \ref{hyperparameters}.

\subsection{Semi-Supervised Results}
\label{sm_results_sec}
For the semi-supervised experiments, report the beam search performance of GRU One2Many for the sake of brevity (greedy performance is in Appendix \ref{more_sm}). In Table \ref{tb:sm_result}, we show the absent performance of GRU One2Many in semi-supervised settings\footnote{Note that \citet{liang-etal-2021-unsupervised} use an extraction model; thus it has no capabilities for absent keyphrase generation}. As we can see from the table, absent performance is near zero in almost all settings. Only when the model is pre-trained (PT) with \textsc{KPDrop-A} or \textsc{KPDrop-R} and then fine-tuned (FT) on LR (PT+KPD-R;FT or PT+KPD-A;FT), there is some degree of absent keyphrase generation. Thus, \textsc{KPDrop} is crucial for downstream absent keyphrase performance in this semi-supervised environment and for domains with low-resource annotated data.

In Table \ref{tb:sm_result}, we also show the present performance of GRU One2Many in semi-supervised settings. For present performance, we find that neither training only (FT) on LR nor training only (PT) on UC is as good as pre-training on UC and then fine-tuning the pre-trained model on LR (PT;FT). Although, one exception is the performance on Inspec which can be sometimes better when the model is zero-shotted after only training on UC. Either way, we again find that using \textsc{KPDrop} in the pre-training (PT+KPD-R;FT or PT+KPD-A;FT) setting significantly boosts present performance after fine-tuning. 

Thus, using \textsc{KPDrop} to make the pre-training stage more challenging by pushing the model to predict missing present keyphrases helps not only with absent keyphrase performance but also with present keyphrase performance after fine-tuning. In between \textsc{KPDrop-R} and \textsc{KPDrop-A}, the former generally performs better in the pre-training stage. Thus, during pre-training it is better to \textit{replace} synthetic labels of fully present keyphrases with its KPDropped version. In Appendix \ref{more_sm}, we also observe similar patterns from other models (One2One and One2Set) in semi-supervised settings. 


\section{Preliminary Experiments on Pre-trained Models}
We also did a few preliminary experiments on applying \textsc{KPDrop-R} to a large pre-trained Seq2Seq language model, in particular, T5 \cite{raffel2020exploring}. We present the results in Appendix \ref{preliminary}. Consistent with previous results, we find that \textsc{KPDrop-R} increases absent performance for T5 as well. However, overall, we found the performance of T5 baseline to be limited compared to trained-from-sratch models like Transformer One2Set. Similarly, other reported performances on pre-trained models have been generally lower than Transformers One2Set too. For example, even after large scale specialized pre-training for keyphrases, \citet{mayank-learning} reports only comparable performance on present keyphrases to Transformer One2Set \cite{ye-etal-2021-one2set} and much less in absent performance. On the other hand, using pre-trained models, \citet{wu-etal-2021-unikeyphrase}\footnote{We are referring to the latest ArXiv version (v2) which holds the globally latest version of the paper.} achieve comparable on absent performance with One2Set under greedy search, but much less in present performance. However, it should not be too difficult to adapt a pre-trained model into a one2set framework during fine-tuning. This can be a promising future direction and form a stronger base model for \textsc{KPDrop}.

\section{Related Work}
\vspace{-2mm}
There is a wide variety of approaches \cite{hasan-ng-2014-automatic,caragea-etal-2014-citation,Das_Gollapalli_Caragea_2014, gollapalli2016keyphrase,sterckx-etal-2016-supervised, florescu-caragea-2017-positionrank, zhang2017mike, boudin-2018-unsupervised, mahata-etal-2018-key2vec, sun2019divgraphpointer,alzaidy2019,campos2020yake, santosh-etal-2020-sasake, Sahrawat2020KeyphraseEA,sun2020joint, song-etal-2021-importance,patel-caragea-2021-exploiting} for keyphrase extraction exclusively. \citet{meng2017deep} diverged from pure extractive methods by introducing a seq2seq model (CopyRNN) for generation of both present and absent keyphrases. \citet{chen-etal-2018-keyphrase} extended CopyRNN with keyphrase correlation constraints and \citet{zhao-zhang-2019-incorporating} extended it with linguistic constraints. \citet{ye-wang-2018-semi, wu-constrained} investigated keyphrase generation (KG) in semi-supervised or resource-constrained settings. \citet{chen2019guided} used a title-guided encoding method for better KG. \citet{wang-etal-2019-topic-aware} incorporated a topic-model to enhance KG. \citet{yuan2020one} extended CopyRNN by introducing the CatSeq model that can generate a concatenation of dynamically determined variable number of keyphrases. \citet{chan-etal-2019-neural, luo2021keyphrase} improved KG using reinforcement learning whereas \citet{Swaminathan_Gupta_Zhang_Mahata_Gosangi_Shah_2020, swaminathan-etal-2020-preliminary, lancioni-etal-2020-keyphrase} do so using GANs. A few approaches \cite{chen-etal-2019-integrated, diao2020keyphrase,kim-etal-2021-structure, ye2021heterogeneous} augmented KG with information from retrieved documents. 

Multiple approaches \cite{chen-etal-2019-integrated, liu2020keyphrase, ahmad2021select,zhao-etal-2021-sgg, wu-etal-2021-unikeyphrase} took a joint-training or multi-tasking approach to do both present keyphrase extraction and absent keyphrase generation. \citet{chen-etal-2020-exclusive} and \citet{ye-etal-2021-one2set} changed the decoder to better respect the structure of keyphrases. \citet{luo2020sensenet} changed the encoder to better respect the input document structure. \citet{Huang_Xu_Jiao_Zu_Zhang_2021} proposed a new beam-search-based adaptive decoding method. \citet{meng-etal-2021-empirical, mayank-learning} investigated pre-training objectives for KG. Both, however, rely on labelled pre-training data. 
\section{Conclusion}
Our proposal, \textsc{KPDrop}, randomly drops present keyphrases from a document to turn them artificially absent. This encourages the model to learn to better exploit the context in the input to be able to infer keyphrases that are absent from the text but otherwise topically relevant. The results show that \textsc{KPDrop} serves as a simple model-agnostic method to substantially improve absent (and sometimes, present) keyphrase performance in both supervised and semi-supervised (low resource) settings when large annotated datasets for keyphrase generation are not available. In future, we would like to explore integration of \textsc{KPDrop} with large-scale pre-training.


\section{Limitations}
\textsc{KPDrop} is a simple yet effective approach to improving performance of keyphrase generation in both large datasets and low resource datasets, which makes it applicable to a wide range of domains where keyphrases are necessary. However, one limitation of \textsc{KPDrop} (especially \textsc{KPDrop-A}) is that it can increase the computation cost during training because both the effective mini-batch size and the effective training dataset size per epoch is doubled through data augmentation. Yet, most data augmentation techniques share the same limitation. 

In addition, \textsc{KPDrop-R} can potentially harm the performance of present keyphrases in some contexts (especially when using greedy search in a supervised setting). To address this, we can simply use the absent predictions of the model trained with \textsc{KPDrop} and the present predictions of the model trained without \textsc{KPDrop}.

\vspace{4mm}
\section{Ethics Statement}
Our technique is specifically designed to improve keyphrase generation. Keyphrase generation is a well-established traditional NLP task that is useful in several application contexts related to organization of information. We do not foresee any immediate ethical concern following from our contribution in this area.

\vspace{5mm}
\section*{Acknowledgements} 
This research is supported in part by NSF CAREER
award \#1802358, NSF CRI award \#1823292, NSF IIS award \#2107518, and UIC Discovery Partners Institute (DPI) award.
Any opinions, findings, and conclusions expressed
here are those of the authors and do not necessarily
reflect the views of NSF or DPI. We thank AWS for computational resources used for this study. We also thank our anonymous
reviewers for their constructive feedback and suggestions.

\bibliography{anthology,custom}
\bibliographystyle{acl_natbib}

\appendix
\newpage\phantom{blabla}\newpage

\begin{table*}[t]
\centering
\small
\def\arraystretch{1.2}
\begin{tabular}{l|cc|cc|cc|cc|cc}
\hline
 & \multicolumn{2}{c}{\textbf{Inspec}} & \multicolumn{2}{|c}{\textbf{NUS}} & \multicolumn{2}{|c}{\textbf{Krapivin}} & \multicolumn{2}{|c}{\textbf{SemEval}} & \multicolumn{2}{|c}{\textbf{KP20k}} \\  \hline
 \multicolumn{1}{c}{\textbf{Models}} & \multicolumn{2}{|c}{\textbf{F1@5C}} & \multicolumn{2}{|c}{\textbf{F1@5C}} & \multicolumn{2}{|c}{\textbf{F1@5C}} & \multicolumn{2}{|c}{\textbf{F1@5C}} & \multicolumn{2}{|c}{\textbf{F1@5C}}\\
 & \textbf{Pre} & {\textbf{Abs}}  & \textbf{Pre} & {\textbf{Abs}} &\textbf{Pre} & {\textbf{Abs}} & \textbf{Pre} & {\textbf{Abs}} & \textbf{Pre} & {\textbf{Abs}} \\ \hline
\multicolumn{11}{l}{\textbf{GRU One2Many}}\\ \hline
Greedy & $\mathbf{22.1_6}$ & $0.7_1$ & $\mathbf{31.2_3}$ & $1.6_3$ & $\mathbf{26.4_5}$ & $1.8_3$ & $\mathbf{24.6_8}$ & $1.6_2$ & $\mathbf{29.3_0}$ & $1.6_0$\\
Greedy+\textsc{KPD-R} & $12.5_3$ & $0.6_0$ & $18.4_1$ & $1.7_3$ & $15.6_3$ & $\mathbf{2.6_1}$ & $13.9_{12}$ & $\mathbf{1.8_1}$ & $17.7_1$ & $1.9_0$ \\
Greedy+\textsc{KPD-A} & $19.7_4$ & $\mathbf{0.8_1}$ & $30.6_6$ & $\mathbf{1.8_3}$ & $24.7_2$ & $2.4_1$ & $23.2_6$ & $\mathbf{1.8_3}$ & $28.1_5$ & $\mathbf{2.0_2}$\\\hdashline
Beam & $\mathbf{34.1_5}$ & $2.3_0$ & $41.3_6$ & $4.0_2$ & $32.6_3$ & $5.2_2$ & $\mathbf{33.3_3}$ & $3.2_1$ & $\mathbf{36.6_0}$ & $4.4_1$\\
Beam+\textsc{KPD-R} & $30.3_2$ & $\mathbf{2.6_2}$ & $37.8_3$ & $\mathbf{5.7_1}$ & $31.9_2$ & $\mathbf{6.6_2}$ & $32.4_{13}$ & $\mathbf{4.3_4}$ & $32.3_1$ & $\mathbf{5.4_0}$\\
Beam+\textsc{KPD-A} & $34.0_5$ & $2.5_1$ & $\mathbf{41.6_7}$ &$5.0_6$ & $\mathbf{33.8_9}$ & $5.5_2$ & $33.1_7$ & $\mathbf{4.3_3}$ & $\mathbf{36.6_0}$ & $5.0_0$ \\ 
\hline
\multicolumn{11}{l}{\textbf{GRU One2One}}\\ \hline
Beam & $30.5_2$ & $2.8_1$ & $40.2_{10}$ & $5.7_7$ & $29.8_1$ & $5.9_3$ & $29.2_{16}$ & $3.8_1$ & $39.4_1$ & $6.2_0$\\
Beam+\textsc{KPD-R} & $30.4_3$ & $\mathbf{2.9_3}$ & $\mathbf{43.9_3}$ & $\mathbf{7.5_3}$ & $\mathbf{36.5_2}$ & $\mathbf{7.8_2}$ & $\mathbf{32.7_{10}}$ & $\mathbf{4.9_5}$ & $37.9_0$ & $6.5_1$\\
Beam+\textsc{KPD-A} & $\mathbf{30.6_2}$ & $2.7_1$ & $38.9_6$ & $6.7_2$ & $29.4_4$ & $6.5_3$ & $30.0_{11}$ & $4.1_3$ & $\mathbf{39.6_0}$ & $\mathbf{6.8_0}$\\
\hline
\multicolumn{11}{l}{\textbf{Transformer One2Set}}\\ \hline
Greedy & $\mathbf{27.6_5}$ & $1.9_3$ & $\mathbf{38.8_4}$ & $4.3_5$ & $\mathbf{30.9_9}$ & $4.1_2$ & $\mathbf{31.2_1}$ & $2.6_2$ & $\mathbf{34.4_3}$ & $3.3_0$\\
Greedy+\textsc{KPD-R} & $15.3_2$ & $2.0_0$ & $26.0_9$ & $\mathbf{5.3_5}$ & $22.2_4$ & $\mathbf{5.9_5}$ & $20.7_7$ & $\mathbf{3.9_3}$ & $24.0_2$ & $\mathbf{4.5_1}$\\
Greedy+\textsc{KPD-A} & $25.7_3$ & $\mathbf{2.1_1}$ & $38.0_2$ & $5.2_8$ & $29.5_7$ & $4.6_4$ & $30.3_7$ & $3.6_1$ & $33.9_3$ & $4.2_1$\\
\hdashline
Beam & $\mathbf{32.3_4}$ & $3.3_2$ & $\mathbf{42.3_3}$ & $7.0_5$ & $32.9_6$ & $6.7_5$ & $33.5_8$ & $4.7_4$ & $\mathbf{36.4_5}$ & $5.8_0$\\
Beam+\textsc{KPD-R} & $27.8_8$ & $2.4_1$ & $40.0_9$ & $7.2_5$ & $\mathbf{33.7_7}$ & $7.3_2$  & $32.0_5$ & $5.2_4$ & $35.1_3$ & $6.1_0$\\
Beam+\textsc{KPD-A} & $32.2_6$ & $\mathbf{3.6_2}$ & $41.8_3$ & $\mathbf{7.9_4}$ & $32.3_7$ & $\mathbf{7.8_2}$ & $\mathbf{34.3_{12}}$ & $\mathbf{5.3_4}$ & $36.3_0$ & $\mathbf{6.7_0}$\\ 
\hline
\end{tabular}
\caption{Present and absent keyphrase performance for different models. Pre represents present performance and Abs represents absent performance. KPD represents KPDrop. Subscripts represent standard deviation (e.g., $31.1_1$ represents $31.1\pm0.1$). We bold the best scores per block.} 
\label{tb:5c_result}
\end{table*}

\begin{table*}[t]
\centering
\small
\def\arraystretch{1.2}
\begin{tabular}{l|cc|cc|cc|cc|cc}
\hline
 & \multicolumn{2}{c}{\textbf{Inspec}} & \multicolumn{2}{|c}{\textbf{NUS}} & \multicolumn{2}{|c}{\textbf{Krapivin}} & \multicolumn{2}{|c}{\textbf{SemEval}} & \multicolumn{2}{|c}{\textbf{KP20k}} \\  
\multicolumn{1}{c|}{\textbf{Models}} & \textbf{F@M} & {\textbf{F1@5}}  & \textbf{F1@M} & {\textbf{F1@5}} & \textbf{F1@M} & {\textbf{F1@5}} & \textbf{F1@M} & {\textbf{F1@5}} & \textbf{F1@M} & \textbf{F1@5} \\ \hline
\multicolumn{11}{l}{\textbf{GRU One2Many Models (Greedy Search)}}\\
\hline
\multicolumn{11}{c}{\textbf{Absent Keyphrase Performance}}\\\hline

PT & $0.0_0$ & $0.0_0$ & $0.0_0$ & $0.0_0$ & $0.0_0$ & $0.0_0$ & $0.0_0$ & $0.0_0$ & $0.0_0$ & $0.0_0$\\
PT+KPD-R & $0.2_1$ & $0.2_1$ & $0.2_2$ & $0.2_2$ & $0.1_0$ & $0.1_0$ & $0.0_0$ & $0.0_0$ & $0.3_1$ & $0.3_1$\\
PT+KPD-A & $0.0_0$ & $0.0_0$ & $0.0_0$ & $0.0_0$ & $0.0_0$ & $0.0_0$ & $0.0_0$ & $0.0_0$ & $0.0_0$ & $0.0_0$\\
FT & $0.1_1$ & $0.1_1$ & $0.0_0$ & $0.1_0$ & $0.2_0$ & $0.2_0$ & $0.3_0$ & $0.3_0$ & $0.1_0$ & $0.1_0$\\
PT; FT & $0.0_1$ & $0.0_1$ & $0.2_1$ & $0.2_1$ & $0.0_0$ & $0.0_0$ & $0.1_1$ & $0.1_1$ & $0.1_0$ & $0.1_0$\\
PT+KPD-A; FT & $0.2_1$ & $0.2_1$ & $\mathbf{0.8_4}$ & $\mathbf{0.8_4}$ & $0.4_1$ & $0.4_1$ & $\mathbf{0.7_0}$ & $\mathbf{0.7_0}$ & $0.6_0$ & $0.6_0$\\
PT+KPD-R; FT & $\mathbf{0.8_3}$ & $\mathbf{0.8_3}$ & $0.5_2$ & $0.5_2$ & $\mathbf{0.5_0}$ & $\mathbf{0.5_0}$ & $0.4_2$ & $0.4_2$ & $\mathbf{0.8_1}$ & $\mathbf{0.8_1}$\\
\hline
\multicolumn{11}{c}{\textbf{Present Keyphrase Performance}}\\\hline
PT & $35.0_7$ & $34.2_6$ & $23.7_2$ & $23.7_2$ & $20.1_4$ & $20.3_2$ & $24.5_6$ & $\mathbf{24.8_5}$ & $18.4_1$ & $18.7_1$\\
PT+KPD-R & $32.0_4$ & $32.0_5$ & $25.3_3$ & $24.8_5$ & $20.5_5$ & $20.3_3$ & $\mathbf{25.0_{11}}$ & $24.7_{11}$ & $19.3_2$ & $19.0_3$ \\
PT+KPD-A & $\mathbf{35.5_5}$ & $\mathbf{34.4_4}$ & $23.8_9$ & $23.9_7$ & $19.3_6$ & $20.0_3$ & $24.5_{15}$ & $24.3_{10}$ & $18.2_4$ & $18.5_3$ \\
FT & $14.3_{12}$ & $14.3_{12}$ & $24.7_{21}$ & $24.7_{21}$ & $23.2_{12}$ & $23.2_{12}$ & $15.8_2$ & $15.8_2$ & $22.8_8$ & $22.8_8$\\
PT; FT & $20.9_{18}$ & $20.9_{18}$ & $27.2_{16}$ & $27.2_{16}$ & $25.5_{10}$ & $25.5_{10}$ & $21.0_5$ & $21.0_5$ & $25.1_7$ & $25.1_7$\\
PT+KPD-A; FT & $22.0_3$ & $22.0_3$ & $\mathbf{30.1_7}$ & $\mathbf{30.1_7}$ & $26.5_7$ & $26.5_7$ & $23.0_6$ & $23.0_6$ & $27.0_2$ & $27.0_2$\\
PT+KPD-R; FT & $20.5_4$ & $20.5_4$ & $\mathbf{30.1_8}$ & $\mathbf{30.1_8}$ & $\mathbf{27.8_4}$ & $\mathbf{27.8_4}$ & $22.3_{12}$ & $22.3_{12}$ & $\mathbf{27.1_2}$ & $\mathbf{27.1_2}$\\
\hline
\end{tabular}%
\vspace{-2mm}
\caption{\small{Absent and present keyphrase performance using Greedy Search for GRU One2Many models in a semi-supervised setup. KPD represents KPDrop. PT represents pre-training on the synthetic data (UC). PT+KPD-R or PT+KPD-A represents pre-training on UC with KPD-R or KPD-A repsectively. FT represents fine tuning or training on the low resource labelled data (LR). PT (or PT+KPD-A or PT+KPD-R) followed by ``; FT" represents that the pre-training was followed by fine-tuning of the pre-trained model on LR. We bold the overall best scores. Subscripts represent standard deviation (e.g., $31.1_1$ represents $31.1\pm0.1$).}} 
\label{tb:sm_greedy_result}
\end{table*}

\begin{table*}[t]
\centering
\small
\def\arraystretch{1.2}
\begin{tabular}{l|cc|cc|cc|cc|cc}
\hline
 & \multicolumn{2}{c}{\textbf{Inspec}} & \multicolumn{2}{|c}{\textbf{NUS}} & \multicolumn{2}{|c}{\textbf{Krapivin}} & \multicolumn{2}{|c}{\textbf{SemEval}} & \multicolumn{2}{|c}{\textbf{KP20k}} \\  
\multicolumn{1}{c|}{\textbf{Models}} & \textbf{F@M} & {\textbf{F1@5}}  & \textbf{F1@M} & {\textbf{F1@5}} & \textbf{F1@M} & {\textbf{F1@5}} & \textbf{F1@M} & {\textbf{F1@5}} & \textbf{F1@M} & \textbf{F1@5} \\ \hline
\multicolumn{11}{l}{\textbf{GRU One2One (Beam Search)}}\\
\hline
\multicolumn{11}{c}{\textbf{Absent Keyphrase Performance}}\\\hline
PT & $0.0_0$ & $0.2_1$ & $0.0_0$ & $0.0_0$ & $0.0_0$ & $0.0_0$ & $0.0_0$ &  $0.1_1$ & $0.0_0$ & $0.1_0$\\ 
PT+KPD-R & $0.4_0$ & $1.2_1$ & $0.7_0$ & $0.9_2$ & $0.6_0$ & $1.6_2$ & $0.9_1$ & $0.9_2$ & $0.4_0$ & $1.3_2$\\
PT+KPD-A & $0.1_0$ & $0.2_1$ & $0.1_0$ & $0.1_1$ & $0.0_0$ & $0.0_0$ & $0.1_0$ & $0.0_0$ & $0.1_0$ & $0.1_0$\\
FT & $0.2_0$ & $0.3_0$ & $0.6_0$ & $1.2_2$ & $0.3_0$ & $0.9_1$ & $0.4_0$ & $0.4_0$ & $0.3_0$ & $0.8_1$\\
PT; FT & $0.2_0$ & $0.6_1$ & $0.6_1$ & $1.3_3$ & $0.4_0$ & $1.0_1$ & $0.3_1$ & $0.5_3$ & $0.3_0$ & $0.8_1$\\
PT+KPD-A; FT & $0.3_0$ & $1.0_1$ & $0.7_0$ & $1.4_2$ & $0.5_0$ & $1.4_1$ & $0.6_1$ & $1.1_3$ & $0.4_0$ & $1.3_0$\\
PT+KPD-R; FT & $\mathbf{0.6_0}$ & $\mathbf{1.9_2}$ & $\mathbf{1.4_0}$ & $\mathbf{4.7_2}$ & $\mathbf{1.1_0}$ & $\mathbf{4.4_2}$ & $\mathbf{1.2_0}$ & $\mathbf{3.0_2}$ & $\mathbf{0.8_0}$ & $\mathbf{3.7_0}$\\
\hline
\multicolumn{11}{c}{\textbf{Present Keyphrase Performance}}\\
\hline
PT & $14.9_1$ & $34.1_8$ & $9.3_0$ & $25.1_4$ & $6.7_0$ & $20.3_6$ & $11.0_1$ & $25.2_3$ & $6.6_1$ & $19.0_1$\\
PT+KPD-R & $\mathbf{21.6_6}$ & $33.6_5$ & $12.6_3$ & $25.3_6$ & $9.8_1$ & $19.8_2$ & $14.6_7$ & $23.1_8$ & $9.2_1$ & $18.4_2$\\
PT+KPD-A & $15.7_2$ & $\mathbf{34.5_5}$ & $9.7_1$ & $25.9_7$ & $7.0_1$ & $20.3_5$ & $11.1_2$ & $25.4_{11}$ & $6.8_1$ & $19.0_5$\\
FT & $20.6_{13}$ & $20.5_{22}$ & $\mathbf{17.8_{15}}$ & $31.9_{20}$ & $\mathbf{12.5_9}$ & $23.4_{17}$ & $\mathbf{16.2_{18}}$ & $21.8_{24}$ & $\mathbf{13.5_8}$ & $25.2_8$\\
PT; FT & $20.6_3$ & $26.1_7$ & $13.7_5$ & $36.6_5$ & $10.1_1$ & $28.2_6$ & $13.9_3$ & $28.1_{15}$ & $10.0_2$ & $29.5_1$\\
PT+KPD-A; FT & $20.6_{10}$ & $27.8_{10}$ & $13.8_8$ & $35.2_{17}$ & $10.1_6$ & $28.9_9$ & $14.0_7$ & $28.7_8$ & $9.9_6$ & $29.2_4$\\
PT+KPD-R; FT & $21.4_3$ & $29.4_8$ & $14.3_2$ & $\mathbf{39.0_8}$ & $10.5_2$ & $\mathbf{30.7_4}$ & $15.0_1$ & $\mathbf{30.4_4}$ & $10.4_2$ & $\mathbf{32.5_1}$\\
\hline
\end{tabular}%
\caption{\small{Absent and present keyphrase performance using Beam Search for GRU One2One models in a semi-supervised setup. KPD represents KPDrop. PT represents pre-training on the synthetic data (UC). PT+KPD-R or PT+KPD-A represents pre-training on UC with KPD-R or KPD-A repsectively. FT represents fine tuning or training on the low resource labelled data (LR). PT (or PT+KPD-A or PT+KPD-R) followed by ``; FT" represents that the pre-training was followed by fine-tuning of the pre-trained model on LR. We bold the overall best scores. Subscripts represent standard deviation (e.g., $31.1_1$ represents $31.1\pm0.1$).}} 
\label{tb:sm_gruone2one_result}
\end{table*}

\begin{table*}[t]
\centering
\small
\def\arraystretch{1.2}
\begin{tabular}{l|cc|cc|cc|cc|cc}
\hline
 & \multicolumn{2}{c}{\textbf{Inspec}} & \multicolumn{2}{|c}{\textbf{NUS}} & \multicolumn{2}{|c}{\textbf{Krapivin}} & \multicolumn{2}{|c}{\textbf{SemEval}} & \multicolumn{2}{|c}{\textbf{KP20k}} \\  
\multicolumn{1}{c|}{\textbf{Models}} & \textbf{F@M} & {\textbf{F1@5}}  & \textbf{F1@M} & {\textbf{F1@5}} & \textbf{F1@M} & {\textbf{F1@5}} & \textbf{F1@M} & {\textbf{F1@5}} & \textbf{F1@M} & \textbf{F1@5} \\ \hline
\multicolumn{11}{l}{\textbf{Transformer One2Set (Greedy Search)}}\\
\hline
\multicolumn{11}{c}{\textbf{Absent Keyphrase Performance}}\\\hline
PT & $0.2_1$ & $0.2_1$ & $0.1_1$ & $0.1_1$ & $0.3_1$ & $0.3_1$ & $0.0_0$ & $0.0_0$ & $0.1_0$ & $0.1_0$\\ 
PT+KPD-R & $0.6_0$ & $0.6_0$ & $0.8_1$ & $0.8_1$ & $0.8_5$ & $0.7_5$ & $0.3_2$ & $0.3_2$ & $0.3_0$ & $0.6_0$\\
PT+KPD-A & $0.3_1$ & $0.3_1$ & $0.2_1$ & $0.2_1$ & $0.2_0$ & $0.2_0$ & $0.1_1$ & $0.1_1$ & $0.2_0$ & $0.2_0$\\
FT & $0.2_3$ & $0.2_3$ & $0.4_3$ & $0.4_3$ & $0.7_3$ & $0.7_3$ & $0.3_1$ & $0.3_1$ & $0.5_1$ & $0.5_1$\\
PT; FT & $0.3_2$ & $0.3_2$ & $0.5_1$ & $0.5_1$ & $0.4_2$ & $0.4_2$ & $0.0_0$ & $0.0_0$ & $0.4_2$ & $0.4_2$\\
PT+KPD-A; FT & $0.5_2$ & $0.5_2$ & $0.7_2$ & $0.7_2$ &$1.1_1$ & $1.1_1$ &$1.1_5$ & $1.1_5$ & $1.2_0$ & $1.2_0$\\
PT+KPD-R; FT & $\mathbf{1.1_1}$ & $\mathbf{1.1_1}$ & $\mathbf{2.9_9}$ & $\mathbf{2.9_9}$ & $\mathbf{2.5_5}$ & $\mathbf{2.5_5}$ & $\mathbf{2.0_6}$ & $\mathbf{2.0_6}$ & $\mathbf{2.4_1}$ & $\mathbf{2.4_1}$\\
\hline
\multicolumn{11}{c}{\textbf{Present Keyphrase Performance}}\\
\hline
PT & $34.6_5$ & $28.6_{17}$ & $25.9_3$ & $23.3_{10}$ & $20.1_4$ & $17.8_4$ & $27.1_4$ & $23.3_7$ & $18.9_1$ & $17.2_4$\\
PT+KPD-R & $\mathbf{35.7_6}$ & $\mathbf{31.7_{13}}$ & $26.3_9$ & $23.9_{11}$ & $20.4_{10}$ & $18.3_5$ & $26.0_7$ & $23.0_{18}$ & $19.3_8$ & $17.8_7$\\
PT+KPD-A & $34.7_9$ & $28.9_5$ & $25.7_4$ & $24.1_{12}$ & $19.8_2$ & $18.5_{10}$ & $26.8_4$ & $23.5_{10}$ & $19.2_1$ & $18.0_{10}$\\
FT & $4.6_7$ & $4.6_7$ & $11.5_{41}$ & $11.5_{41}$ & $9.5_{11}$ & $9.5_{11}$ & $6.1_{17}$ & $6.1_{17}$ & $8.5_{24}$ & $8.5_{24}$\\
PT; FT & $20.6_{14}$ & $20.6_{14}$ & $29.2_{20}$ & $29.2_{20}$ & $24.9_{11}$ & $24.9_{11}$ & $23.2_{16}$ & $23.2_{16}$ & $24.7_{15}$ & $24.7_{15}$\\
PT+KPD-A; FT & $23.3_5$ & $23.3_5$ & $32.5_{10}$ & $32.3_9$ & $27.1_9$ &  $27.1_8$ & $26.7_{11}$ & $26.7_{12}$ & $27.5_2$ & $27.4_2$\\
PT+KPD-R; FT & $26.4_{19}$ & $26.4_{19}$ & $\mathbf{37.0_{10}}$ & $\mathbf{36.6_8}$ & $\mathbf{30.6_{10}}$ & $\mathbf{30.3_8}$ & $\mathbf{28.9_{14}}$ & $\mathbf{28.5_{12}}$ & $\mathbf{30.9_5}$ & $\mathbf{30.6_4}$\\
\hline
\end{tabular}%
\caption{\small{Absent and present keyphrase performance using Greedy Search for Transformer One2Set models in a semi-supervised setup. KPD represents KPDrop. PT represents pre-training on the synthetic data (UC). PT+KPD-R or PT+KPD-A represents pre-training on UC with KPD-R or KPD-A repsectively. FT represents fine tuning or training on the low resource labelled data (LR). PT (or PT+KPD-A or PT+KPD-R) followed by ``; FT" represents that the pre-training was followed by fine-tuning of the pre-trained model on LR. We bold the overall best scores. Subscripts represent standard deviation (e.g., $31.1_1$ represents $31.1\pm0.1$).}} 
\label{tb:sm_trans2set_greedy_result}
\end{table*}

\begin{table*}[t]
\centering
\small
\def\arraystretch{1.2}
\begin{tabular}{l|cc|cc|cc|cc|cc}
\hline
 & \multicolumn{2}{c}{\textbf{Inspec}} & \multicolumn{2}{|c}{\textbf{NUS}} & \multicolumn{2}{|c}{\textbf{Krapivin}} & \multicolumn{2}{|c}{\textbf{SemEval}} & \multicolumn{2}{|c}{\textbf{KP20k}} \\  
\multicolumn{1}{c|}{\textbf{Models}} & \textbf{F@M} & {\textbf{F1@5}}  & \textbf{F1@M} & {\textbf{F1@5}} & \textbf{F1@M} & {\textbf{F1@5}} & \textbf{F1@M} & {\textbf{F1@5}} & \textbf{F1@M} & \textbf{F1@5} \\ \hline
\multicolumn{11}{l}{\textbf{Transformer One2Set (Beam Search)}}\\
\hline
\multicolumn{11}{c}{\textbf{Absent Keyphrase Performance}}\\\hline
PT & $0.2_0$ & $0.7_2$ & $0.2_0$ & $0.6_3$ & $0.1_0$ & $0.2_1$ & $0.2_0$ & $0.2_2$ & $0.1_0$ & $0.3_0$\\ 
PT+KPD-R & $\mathbf{0.4_0}$ & $1.4_2$ & $0.5_0$ & $1.4_2$ & $0.4_0$ & $1.2_{12}$ & $0.5_1$ & $1.2_2$ & $0.3_0$ & $1.2_1$\\
PT+KPD-A & $0.3_0$ & $0.8_1$ & $0.3_0$ & $0.4_0$ & $0.2_0$ & $0.3_1$ & $0.2_0$ & $0.3_1$ & $0.1_0$ & $0.4_0$\\
FT & $0.2_0$ & $0.3_2$ & $0.6_0$ & $0.8_2$ & $0.3_0$ & $1.0_2$ & $0.4_1$ & $0.8_1$ & $0.3_0$ & $0.9_0$\\
PT; FT & $0.3_0$ & $0.8_1$ & $0.5_0$ & $1.6_3$ & $0.3_0$ & $1.4_1$ & $0.4_0$ & $0.6_1$ & $0.3_0$ & $1.1_1$\\
PT+KPD-A; FT & $0.3_0$ & $1.0_3$ & $0.5_0$ & $1.8_4$ & $0.4_0$ & $1.7_3$ & $0.4_0$ & $1.5_2$ & $0.3_0$ & $1.7_0$\\
PT+KPD-R; FT & $\mathbf{0.4_0}$ & $\mathbf{1.7_1}$ & $\mathbf{1.0_0}$ & $\mathbf{5.4_4}$ & $\mathbf{0.7_0}$ & $\mathbf{4.5_2}$ & $\mathbf{0.9_0}$ & $\mathbf{3.1_2}$ & $\mathbf{0.5_0}$ & $\mathbf{3.6_0}$\\
\hline
\multicolumn{11}{c}{\textbf{Present Keyphrase Performance}}\\
\hline
PT & $19.3_8$ & $28.6_{15}$ & $16.2_3$ & $22.7_9$ & $10.3_3$ & $17.5_4$ & $18.0_9$ & $23.4_{11}$ & $10.5_4$ & $17.0_5$\\
PT+KPD-R & $\mathbf{22.5_4}$ & $\mathbf{31.9_{15}}$ & $16.9_9$ & $23.6_7$ & $\mathbf{11.3_5}$ & $18.0_6$ & $\mathbf{19.6_{10}}$ & $22.9_{15}$ & $10.8_5$ & $17.6_7$\\
PT+KPD-A & $20.7_3$ & $28.8_8$ & $\mathbf{17.2_4}$ & $23.5_9$ & $11.1_2$ & $17.9_{10}$ & $19.0_5$ & $22.9_{15}$ & $\mathbf{11.3_2}$ & $17.8_9$\\
FT & $12.6_{10}$ & $11.3_5$ & $15.0_{15}$ & $23.7_9$ & $9.3_9$ & $14.2_7$ & $12.3_{10}$ & $15.3_{11}$ & $10.9_8$ & $16.8_1$\\
PT; FT & $17.4_2$ & $24.3_8$ & $14.9_3$ & $32.1_6$ & $9.5_{12}$ & $25.0_2$ & $14.8_5$ & $27.4_{11}$ & $10.5_2$ & $25.6_3$\\
PT+KPD-A; FT & $17.7_3$ & $25.9_2$ & $15.7_1$ & $33.2_{11}$ & $10.1_2$ & $25.8_5$ & $15.8_2$ & $26.7_{12}$ & $10.9_1$ & $26.8_2$\\
PT+KPD-R; FT & $20.4_4$ & $\mathbf{31.9_9}$ & $15.4_2$ & $\mathbf{37.1_5}$ & $10.3_2$ & $\mathbf{29.2_7}$ & $16.6_2$ & $\mathbf{30.4_{18}}$ & $10.2_2$ & $\mathbf{29.8_1}$\\
\hline
\end{tabular}%
\caption{\small{Absent and present keyphrase performance using Beam Search for Transformer One2Set models in a semi-supervised setup. KPD represents KPDrop. PT represents pre-training on the synthetic data (UC). PT+KPD-R or PT+KPD-A represents pre-training on UC with KPD-R or KPD-A repsectively. FT represents fine tuning or training on the low resource labelled data (LR). PT (or PT+KPD-A or PT+KPD-R) followed by ``; FT" represents that the pre-training was followed by fine-tuning of the pre-trained model on LR. We bold the overall best scores. Subscripts represent standard deviation (e.g., $31.1_1$ represents $31.1\pm0.1$).}} 
\label{tb:sm_trans2set_beam_result}
\end{table*}

\section{Appendix}
\label{sec:appendix}

\subsection{Hyperparameters}
\label{hyperparameters}
For GRU One2Many (CatSeq) we use hyperparameters similar to prior works \cite{chan-etal-2019-neural}. Like prior works \cite{yuan2020one, meng-etal-2021-empirical}, when using beam search on one2many models we use a beam size of $50$. We use the same model hyperparameters of GRU One2One models as used for GRU One2Many. However, during beam search we use a beam size of $200$ as is standard for one2one models in prior work \cite{meng2017deep, meng-etal-2021-empirical}. A lower beam size ($50$ instead of $200$) is used for one2many models because they can generate multiple keyphrases per beam. So a lower size is used to make them more comparable with one2one models with higher beam size. For Transformer One2Set model we use the same hyperparameters as \citet{ye-etal-2021-one2set}. Given the similarity of One2Set models with One2Many models in their ability to generate multiple keyphrases per beam, we also use a beam size of $50$ for for Transformer One2Set. Anytime we use beam search, we also use length normalization on beam search with a length co-efficient of $0.8$. We use the same settings during semi-supervised pre-training or fine-tuning. We tested \textsc{KPDrop-R} rates among \{$0.3, 0.5, 0.7, 0.9, 1.0$\} on GRU One2Many over the validation set after one epoch training on the full KP20K training set. We found $0.7$ to be the best performing rate for validation absent performance. We use this same rate for \textsc{KPDrop-A} and other models in both supervised and semi-supervised settings. All the experiments were done in a single Nvidia RTX A6000.

\subsection{More Evaluations}
\label{more_evaluation}
\citet{chan-etal-2019-neural} modified the original F$_1$@5 (as used in the main paper and other prior works \cite{meng2017deep, yuan2020one})  such that the denominator in the precision is always set to $5$ even when the total predictions are less than $5$.  To avoid confusion and better distinguish from the original F$_1$@5 we refer to the modified metric as F$_1$@5C. Throughout the paper, for the sake of brevity we mainly report performance in F$_1$@5 instead of F$_1$@5C. This is because although F$_1$@5C achieve the goal of differentiating itself from F$_1$@M reports in greedy search contexts, it can be a little misleading. For example, F$_1$@5C can artificially penalize the model for predicting less than $5$ keyphrases even when the ground truth itself contains less than $5$ keyphrases. Nevertheless, in Table \ref{tb:5c_result}, we still report the F$_1$@5C present and absent performance of our models from our supervised experiments for the sake of better comparison with prior works that use F$_1$@5C. 

\subsection{More Semi-Supervised Experiments}
\label{more_sm}
In Table \ref{tb:sm_greedy_result} we present the greedy decoding performance of GRU One2Many models in semi-supervised settings. In Table \ref{tb:sm_gruone2one_result} we present the performance of GRU One2One models (beam search) in semi-supervised settings. In Table \ref{tb:sm_trans2set_greedy_result} and \ref{tb:sm_trans2set_beam_result} we present the greedy decoding performance and beam decoding performance of Transformer One2Set mdoels in semi-supervised settings respectively. Gnerally, we notice similar patterns across all the models as were found and discussed for GRU One2Many models in $\S$\ref{sm_results_sec}. Overall, \textsc{KPDrop-R} consistently serves as a crucial ingredient in the pre-training stage to enable substantially improved downstream performance in both present and absent keyphrase generation.

\subsection{Preliminary Experiments on Additional Baselines}
\label{preliminary}
We also present the results of applying \textsc{KPDrop-R} on a large pre-trained model T5 \cite{raffel2020exploring}, and another specialized KG model, ExHiRD \cite{chen-etal-2020-exclusive} in Table \ref{tb:aux_absent_result}. In both cases, we see the promise of  \textsc{KPDrop-R} in improving the absent performance. 

Below we describe the hyperparameters that were used for the preliminary experiments described in this section. 

For ExHiRD, we use the same hyperparameters as in the original paper \cite{chen-etal-2020-exclusive}\footnote{\url{https://github.com/Chen-Wang-CUHK/ExHiRD-DKG}}. For T5, we use a maximum of $10$ epochs, early stopping with a patience of $2$ (the training is terminated if validation loss does not improve for $2$ consecutive epochs), a batch size as $64$, a maximum gradient norm clipping af $5$, and SM3 \cite{anil2019memory}  as the optimizer.  The initial learning rate for T5 was set to be $0.1$. We apply learning rate (lr) warmup  as follows \footnote{based on the recommended procedure for SM3 (\url{https://github.com/google-research/google-research/tree/master/sm3}).}:
\begin{equation}
    lr_s = lr_0 \cdot min(1,(s/w)^2)
\end{equation}
$lr_s$ indicates the learning rate at step $s$. $lr_0$ is the initial learning rate ($0.1$). $s$ indicates the current update step number. $w$ indicates total warmup steps (set as $2000$). The initial learning rate ($0.1$) was tuned using grid search based on validation loss among the following choices: $\{0.1, 0.01, 0.001\}$. For each trial during hyperparameter optimization we use a maximum of $1$ epoch. We only tune the baselines (where \textsc{KPDrop} is unapplied). We do not separately tune other hyperparameters when \textsc{KPDrop} is applied. 

We use the T5 implementation as provided by \cite{wolf2020transformers}. Both ExHiRD and T5 models are trained using teacher forcing mechanism.  During inference, we set the maximum phrase length as $50$ for both. Keyphrase tokens are greedily generated during inference. A \textsc{KPDrop} rate of $0.7$ was used - same as the other previous experiments. 
\begin{table*}[t]
\centering
\small
\def\arraystretch{1.2}
\begin{tabular}{l|cc|cc|cc|cc}
\hline
 & \multicolumn{2}{c}{\textbf{Inspec}} & \multicolumn{2}{|c}{\textbf{Krapivin}} & \multicolumn{2}{|c}{\textbf{SemEval}} & \multicolumn{2}{|c}{\textbf{KP20k}} \\  
\multicolumn{1}{c|}{\textbf{Models}} & \textbf{F1@M} & {\textbf{F1@5C}} & \textbf{F1@M} & {\textbf{F1@5C}} & \textbf{F1@M} & {\textbf{F1@5C}} & \textbf{F1@M} & \textbf{F1@5C} \\ \hline
ExHiRD Greedy & $2.2$ & $1.1$ & $4.3$ & $2.2$& $2.5$ & $1.7$ & $3.2$ & $1.6$\\
ExHiRD Greedy+KPD-R & $\mathbf{3.5}$ & $\mathbf{2.0}$ & $\mathbf{6.8}$ & $\mathbf{3.7}$ & $\mathbf{5.1}$ & $\mathbf{3.7}$ & $\mathbf{5.3}$ & $\mathbf{2.7}$\\
\hline
T5 Greedy & $2.5$ & $1.4$ & $5.3$ & $2.8$ & $2.3$ & $1.6$ & $3.6$ & $1.8$\\
T5 Greedy+KPD-R & $\mathbf{3.2}$ & $\mathbf{1.8}$ & $\mathbf{7.0}$ & $\mathbf{4.2}$ & $\mathbf{3.8}$ & $\mathbf{2.9}$ & $\mathbf{5.7}$ & $\mathbf{3.1}$ \\
\hline
\end{tabular}%
\vspace{-2mm}
\caption{\small{Absent keyphrase performance of ExHiRD and T5. We bold the best scores per block.}} 
\label{tb:aux_absent_result}
\end{table*}

\end{document}